\pdfoutput=1
\documentclass[10pt,twocolumn,letterpaper]{article}

\usepackage{iccv}
\usepackage{times}
\usepackage{epsfig}
\usepackage{graphicx}
\usepackage{amsmath}
\usepackage{amssymb}
\usepackage{kotex}
\usepackage{subfigure}
\usepackage{multirow}

\usepackage[breaklinks=true,bookmarks=false]{hyperref}
\iccvfinalcopy % *** Uncomment this line for the final submission

\def\iccvPaperID{9684} % *** Enter the ICCV Paper ID here

\ificcvfinal\fi

\begin{document}

%%%%%%%%% TITLE
\title{\ Gesture Recognition with a Skeleton-Based Keyframe Selection Module}

\author{Yunsoo Kim\\
School of Electrical Engineering, KAIST\\
Daejeon, 305-701, Republic of Korea\\
{\tt\small dbstn1121@kaist.ac.kr}
\and
Hyun Myung\\
School of Electrical Engineering, KAIST\\
Daejeon, 305-701, Republic of Korea\\
{\tt\small hmyung@kaist.ac.kr}
}

\maketitle
\ificcvfinal\thispagestyle{empty}\fi

%%%%%%%%% ABSTRACT
\begin{abstract}
We propose a bidirectional consecutively connected two-pathway network (BCCN) for efficient gesture recognition. The BCCN consists of two pathways: ($i$) a keyframe pathway and ($ii$) a temporal-attention pathway. The keyframe pathway is configured using the skeleton-based keyframe selection module. Keyframes pass through the pathway to extract the spatial feature of itself, and the temporal-attention pathway extracts temporal semantics. Our model improved gesture recognition performance in videos and obtained better activation maps for spatial and temporal properties. Tests were performed on the Chalearn dataset, the ETRI-Activity 3D dataset, and the Toyota Smart Home dataset.
\end{abstract}

%%%%%%%%% BODY TEXT
\section{Introduction}

The purpose of gesture recognition is to distinguish users' gestures in videos. Gesture recognition is given a video sequence as input, unlike an image in ImageNet \cite{deng2009imagenet}. Thus, simply using spatial features of video frames cannot effectively achieve gesture recognition. Spatiotemporal networks exploit temporal features as well as spatial features for gesture recognition \cite{feichtenhofer2017spatiotemporal, taylor2010convolutional, tran2015learning}. Furthermore, multi-modal networks leverage a variety of modalities to improve the network to contain more information \cite{nishida2015multimodal, zhu2017multimodal}. Also, studies have been conducted using spatiotemporal features drawn from several modalities at once \cite{neverova2015moddrop}.

\begin{figure*}[t]
    \centerline{\includegraphics[width=0.98\textwidth]{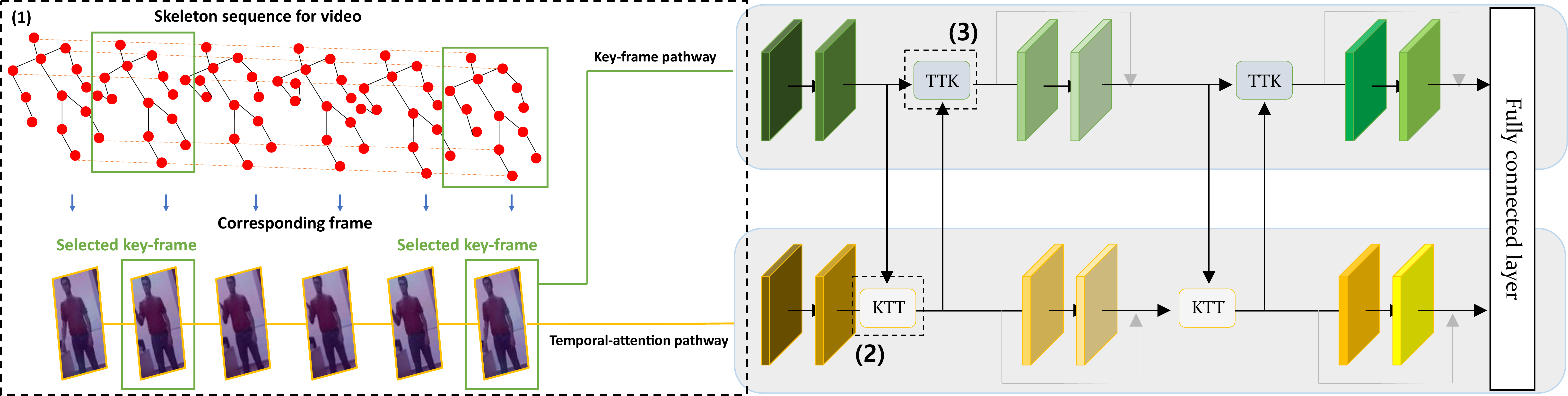}}
    \caption{An overview of the proposed network architecture for efficient gesture recognition. The inputs are video frames and skeleton data, and the output is the predicted class of gestures in the video. Our network can be largely divided into three stages. (1) Extract the keyframes from the video using skeleton sequences. (2) Feed the keyframe features from the keyframe pathway to the temporal-attention pathway. (3) Extract the spatiotemporal features from the temporal-attention features and join it into the keyframe pathway. Step (2) through (3) operate continuously and repeatedly.
    } \label{1}
\end{figure*}

Despite utilizing temporal features, existing studies have limitations in extracting efficient spatiotemporal features. Based on this intuition, we attempt to extract efficient temporal features from video sequences with skeleton sequences. Skeleton sequences and video sequences are deeply related to each other, and they are characterized by sharing time-critical points. We propose a skeleton-based keyframe selection module to extract the critical time points from the video data and propose a network that can maintain the keyframe features and the temporal-attention features well, as can be found in Fig. \ref{1}.

There are also studies that use different modalities to efficiently extract temporal features. Gao \textit{et al.} \cite{gao2020listen} conducted a study using audio sequences to find important parts within video clips which contain corresponding audio. They can efficiently select critical moments within a long untrimmed video to extract time-critical features. In addition, Song \textit{et al.} \cite{song2020gesture} uses audio sequences as a detector for gestures. The detector that uses audio data determines which part of the video to proceed with for gesture recognition. Likewise, we do not simply fuse modalities \cite{miao2017multimodal}, but exploit skeleton sequences to extract keyframes within video sequences.

We leverage attention-based long short-term memory (LSTM) to select the keyframes of the skeleton sequences, and use the frames as the keyframes of the video sequences. Furthermore, we prove that utilizing skeleton data as a keyframe selection module can achieve higher performance than simply fusing the modalities. We also propose a bi-directional consecutively connected two-pathway network (BCCN) that efficiently fuses keyframe features and spatiotemporal features. By keeping the keyframe features in the two pathways, both pathways are able to extract effective features for gesture recognition, and the activation maps are also obtained more clearly.

Our contributions can be summarized as follows:
\begin{itemize}
\item We experimentally show how the skeleton-based keyframe selection module helps achieve good keyframe features for gesture recognition.
\item We propose a BCCN that can contain keyframe features and spatiotemporal features well during the learning process to classify gestures efficiently.
\item The pathways of the proposed network teach each other to maintain spatiotemporal characteristics of both pathways for fusing spatial semantics and temporal semantics effectively.
\end{itemize}

\section{Related Work}
In this section, we consider the following topics: temporal semantics, multimodal network, multitask learning, skeleton data, and two-stream networks.

\noindent\textbf{Temporal Semantics}
Gestures consist of a series of human poses, so they cannot be thought of without temporal semantics. Therefore, it is common to use video sequences for gesture recognition. However, video sequences contain many parts that are unnecessary to recognize the gestures. For example, video sequences can include basic human postures that are not related to behavior or contain the movements of objects other than humans. In addition, adjacent frames are likely to have many overlapping features because human body positions are likely to be similar to each other. In this cases, the network learns unnecessary information, which leads to poor learning performance. Therefore, efficient handling of temporal semantics is considered an important issue for improving performance.

Neverova \textit{et al.} \cite{neverova2015moddrop} uses various time scales to process temporal information, helping to perceive behaviors insensitive to the speed of the gestures. However, it requires diversifying the time scales constantly to get better performance. In comparison, Feichtenhofer \textit{et al.} \cite{feichtenhofer2019slowfast} proposed a method to follow a two-pathway network. The two pathways that make up the network extract features, focusing on the spatial and temporal characteristics, respectively. Unlike Neverova's work, it is efficient in that it extracts specific features separately -- the temporal features of the video sequences and spatial features of the frames. However, this network can be improved because it only extracts spatial features of the starting frames from each video batch.

\noindent\textbf{Multimodal Networks}
In addition to simply handling time information well, there are ways to increase gesture recognition performance by efficiently combining and accepting different kinds of data. For example, a person does not only use their vision to grasp the behavior of others. For example, when you see a person waving their hand forward or backward, it is efficient to use other senses, such as hearing, to determine whether the person is asking you to come or to go. Even if a person is doing the same thing, when they say ``Come this way,'' the action becomes ``come this way.'' But when they say ``Go that way,'' the action becomes ``go that way.'' Therefore, the multimodal network takes the approach that multiple behaviors can be classified well if other senses are available.

Multimodal networks do not use one type of data, but rather two or more types of data. Multimodal networks increase behavioral recognition performance, but due to the large size of the data, the performance of the hardware is greatly affected. This is why the way to combine different types of data is important. The modality fusion method varies the timing and feature combining method to handle the data efficiently \cite{gadzicki2020early, miao2017multimodal, snoek2005early, wang2020deep}. Recently, a method has been proposed to effectively extract features from the data using other types of data \cite{gao2020listen, song2020gesture}. We discuss in Section 4 that it is better to extract effective features from other data using one type of data rather than simple fusion methods in the field of gesture recognition.

\noindent\textbf{Multitask Learning}
Multitask learning is a learning method that improves the performance of all tasks by allowing a system to learn highly correlated tasks simultaneously \cite{caruana1997multitask, crawshaw2020multi, ruder2017overview}. The network can be learned efficiently because it shares learned representations, which helps to obtain generalized features. Various sharing methods are being studied \cite{kumar2012learning, meyerson2017beyond}. There are three issues in multitask learning: when to share, what to share, and how to share. Various studies have been conducted to tackle these issues \cite{argyriou2008convex, han2016multi, liu2015multi}.

Luvizon \textit{et al.} \cite{luvizon20182d} utilizes a multitask CNN to perform simultaneous appearance and pose recognition, combining them for action recognition. By performing similar tasks using one network, the CNN was able to learn the generalized shared presentation well. Likewise, we focus on extracting generalized features while obtaining the spatiotemporal features.

\noindent\textbf{Skeleton Data}
Skeleton data contains information about human skeletons and joints, which is important in determining human gestures. Skeleton data can be extracted using RGB-D sensors, which consist of cameras and infrared sensors, or using deep learning \cite{fang2017rmpe, kim2015real, toshev2014deeppose}, such as OpenPose \cite{cao2019openpose} or AlphaPose \cite{xiu2018pose}. As such, skeleton data are basically extracted from image frames. This is why image sequences and skeleton sequences have similar time features.

There are some studies on recognizing gestures with skeleton data only \cite{kim2016weighted, shi2019skeleton}. Shi \textit{et al.} \cite{shi2019skeleton} proposed a method of sequentially applying spatial attention modules and temporal attention modules to extract features to proceed with gesture recognition using selected features. They have shown that gesture recognition is possible with skeleton data alone, but there are limitations in distinguishing similar behaviors; that's why we decided to utilize skeleton data to select keyframes of gestures.

\noindent\textbf{Two-Stream Network}
Multimodal networks use multistream networks that process each modality separately. The network can be learned efficiently because each modality has different characteristics and each type of data can fit efficiently. In addition, the pathways can be divided into spatial and temporal pathways. In the spatial pathway, they focus on the spatial features of objects that can be drawn from each frame, and in the temporal pathway, they focus on the temporal features of motion through the frames.

To fuse the processed features effectively, we use a lateral connection. Feichtenhofer \textit{et al.} \cite{feichtenhofer2016convolutional} proposed a way to fuse optical flow with video sequences to achieve better gesture recognition results. Through a lateral connection and fusion between the pathways, information about the optical flow and video sequences is fused, helping the network distinguish the gestures well. We constructed a network with two pathways -- the keyframe pathway and temporal-attention pathway. We also propose the keyframe to temporal attention (KTT) unit and the temporal attention to keyframe (TTK) unit for efficient exchanges of information, which are discussed in Section 3.

\section{The Proposed Method}
Our goal is to achieve efficient gesture recognition using skeleton and video data. We first propose a module using skeleton data to pull the keyframe out of the video (Section 3.1). Also, we present a network that can effectively deliver the keyframe feature (Section 3.2).

\subsection{Skeleton-Based Keyframe Selection Module}
\begin{figure}[t]
    \centerline{\includegraphics[width=0.49\textwidth]{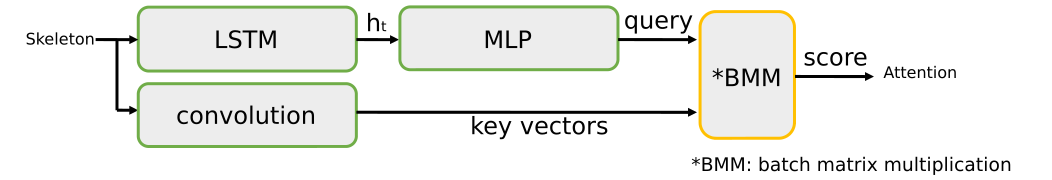}}
    \caption{Structure of the keyframe selection module using skeleton data.
    } \label{2}
\end{figure}

As can be found in Fig. \ref{2}, the keyframe selection module uses long short-term memory and attention mechanisms to extract keyframes. First, the skeleton sequence enters into the long short-term memory. Each output continues to update the hidden layer of the long short-term memory, and the updated hidden value ($h_t$) passes through the multilayer perceptron to form a query vector. We formed a key vector through $1 \times 1$ convolution for each skeleton sequence. The attention score is obtained by multiplying the matrix of query vectors with key vectors. The keyframe of skeleton sequences is selected by the softmax function and the maximum value function of the attention score.

Two criteria must be satisfied to utilize the skeleton-based keyframe selection module. 1) Skeleton data contains sufficient data for action recognition. 2) Keyframes in skeleton data and video sequences should be similar. For the first prerequisite, as demonstrated by Shi \textit{et al.} \cite{shi2019skeleton}, our network contains sufficient data for behavior recognition, since behavior recognition is also possible only with skeleton data. Experiments conducted to prove the second prerequisite are discussed in Section 4.1.

\subsection{Proposed Network}
The goal of a BCCN is that the two pathways that make up the network can learn from each other to better focus on the regions needed to recognize gestures. Rather than simply drawing out temporal and spatial features, each characteristic complements the other to achieve better gesture recognition results.

\begin{figure*}[t]
    \centerline{\includegraphics[width=.98\textwidth]{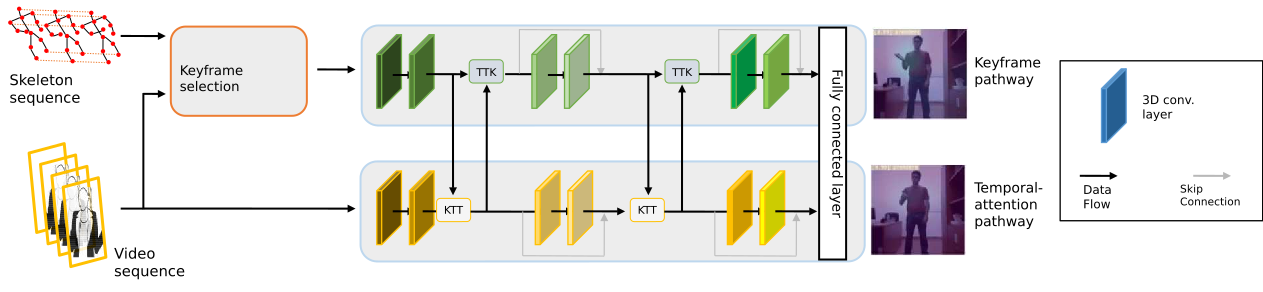}}
    \caption{The architecture of the proposed network BCCN. The BCCN consists of three units: skeleton-based keyframe selection module, keyframe to temporal attention (KTT) unit, and temporal attention to keyframe (TTK) unit. The units help the network to maintain efficient spatiotemporal features.
    } \label{3}
\end{figure*}

\noindent\textbf{BCCN}
The architecture of the proposed BCCN can be found in Fig. \ref{3}. It has a two-pathway network structure connected continuously in lateral directions. The first pathway contains keyframes selected by the previously proposed keyframe selection module, and the second pathway contains general video sequences. Each feature is learned from a 3D convolution layer in each pathway, and the network's structure provides features to each pathway by sequentially applying the KTT unit and the TTK unit.

The features that can be obtained from the first pathway are mainly focused on spatial characteristics. Since the first pathway contains the keyframe extracted from the video, we extract the keyframe spatial feature on the basis that it contains a lot of important information in gesture recognition. The extracted keyframe features are provided to the second path via the KTT unit to help the network determine gestures by weighting the keyframe features. In the second pathway, the number of frames entering the input is increased, and extracted features are more focused on temporal semantics. The extracted temporal-attention features are supplied to the first pathway through the TTK unit. The TTK unit also considers spatial aspects of the temporal-attention features that are less dependent on the size of the features.

\begin{figure}[t]
    \centerline{\includegraphics[width=0.49\textwidth]{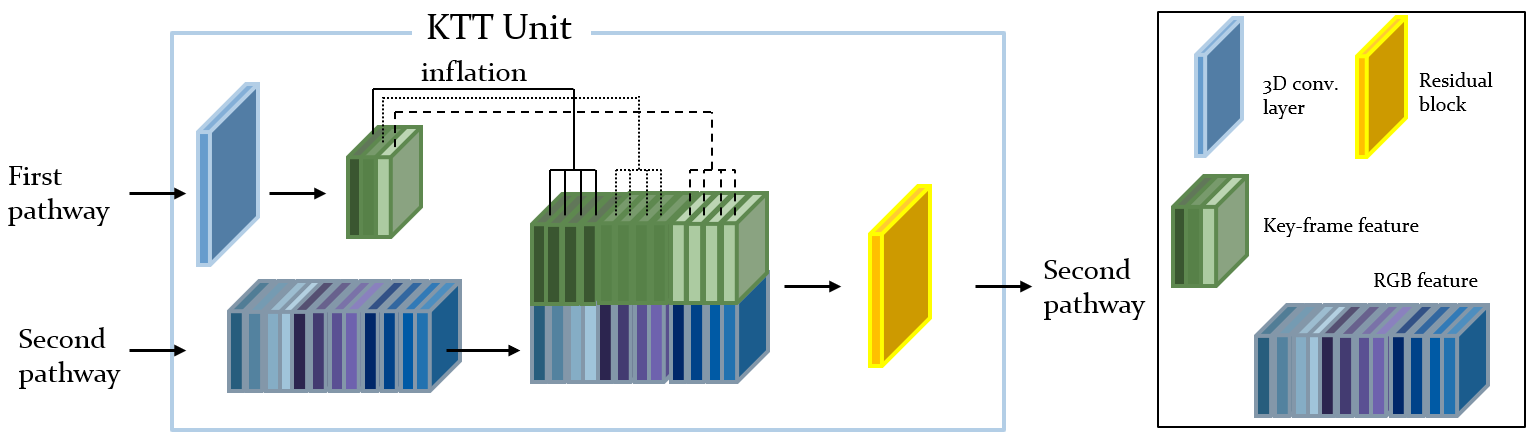}}
    \caption{The architecture of the keyframe to temporal attention (KTT) unit. The unit inflates the keyframe features to give weights to the features through the network.
    } \label{4}
\end{figure}

\noindent\textbf{KTT unit}
The KTT unit, as can be found in Fig. \ref{4}, helps to focus better on keyframe features within the network. The unit inflates the keyframe features, which are selected through the keyframe selection module, and supplies them to the second pathway. Basically, the keyframe features in the first pathway have a small dimension on the time axis, instead of a large channel, and contain a variety of spatial information. Therefore, to supply the keyframe features from the first pathway to the second pathway, the channel must be reduced and amplified over time. To this end, the time step must be unified by aligning the channels through the 3D convolution layer and then inflating over time.

By adding keyframe features for the time step that contains each of the temporal-attention features, the influence of the keyframe features can be enhanced. It is necessary to focus more on the spatial features to carefully distinguish similar behaviors. Therefore, strengthening the keyframe features through KTT units serves to better distinguish similar behaviors. The second pathway, enhanced through the KTT unit, is followed by the TTK unit.

\begin{figure}[t]
    \centerline{\includegraphics[width=0.49\textwidth]{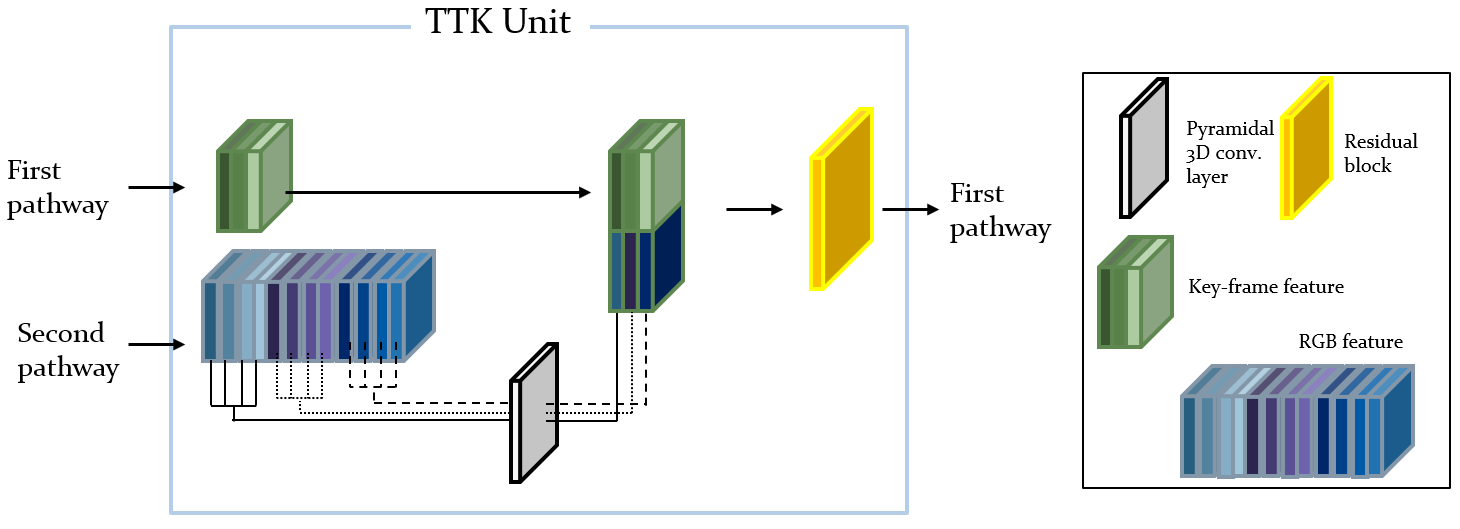}}
    \caption{The architecture of the temporal attention to keyframe (TTK) unit. The unit applies the pyramidal 3D convolution layer to extract size-insensitive sptiotemporal features.
    } \label{5}
\end{figure}

\noindent\textbf{TTK unit}
The TTK unit, as can be found in Fig. \ref{5}, serves to reinforce the spatial features by extracting spatiotemporal information from the temporal-attention features drawn from the second pathway and feeding them to the first pathway. Because the temporal-attention features in the second pathway have focused on temporal information, instead of having a large dimension on the time axis, temporal-attention features have fewer channels. Therefore, to supply the spatiotemporal features from the second pathway to the first pathway, the channel must be increased and compressed over time. For this purpose, the channel and the time dimension must be matched through the pyramidal 3D convolution layer and supplied to the first pathway according to each time step.

The time dimension of the keyframe features can also be aligned with the time dimension of the temporal-attention features, which flow along the second pathway, by using a basic 3D convolution layer. However, the reason for applying the pyramidal 3D convolution layer is to make the spatiotemporal features insensitive to features' size when being extracted. For good gesture recognition, all of the features must be well recognized regardless of their size. In other words, it should work regardless of size of the behavior or objects. To this end, varying the size of the kernels results in varying the receptive field, which results in varying the size of the regions in which the spatiotemporal features are extracted. In other words, the spatiotemporal features, which are insensitive to differences in size through the TTK unit, improve the performance of gesture recognition.

The features from the two pathways that pass the KTT unit and the TTK unit know where to look through each other's information. In particular, since the second pathway develops from the existing temporal-attention features, it is possible to focus on the important part and extract better temporal features for that part.

\section{Experiments}
\noindent\textbf{Datasets}
We conducted training on three main datasets: Chalearn \cite{escalera2014chalearn}, ETRI-Activity3D \cite{jang2020etri}, and Toyota Smarthome \cite{das2019toyota}. Each dataset contains Italian gestures, gestures in the indoor environment of apartment, and gestures in the real-life environment of the elderly, respectively. We used all of the Chalearn dataset and 14 classes from ETRI-Activity3D and Toyota Smarthome datasets. Each dataset contains different age groups, so the speed of the actions is different, and each video has different duration. It is important to select spatiotemporal features efficiently to distinguish gestures accordingly. Using the video and skeleton data present in each dataset, we conducted the following experiments. If a video contained multiple gestures, the experiments were conducted by separating them and dividing them into multiple video clips.

\noindent\textbf{Implementation Details}
We conducted an experiment using PyTorch. For fast learning, we resized the frame inputs for each image sequence to 178 $\times$ 120. For skeleton data, experiments were conducted using only the \textit{x}-axis, \textit{y}-axis, and \textit{z}-axis values for each joint. The skeleton-based keyframe selection module used 1,024 hidden units and 512 key dimensions. We trained the network for 150 epochs. We used momentum of 0.9 and weight decay of $10^{-4}$.

\noindent\textbf{Evaluation}
We measured the recognition accuracy for each gesture for evaluation. If the top-predicted target was the same as the actual label, it was determined that it is the correct prediction. We also checked the activation map whether the part that should be activated to determine the actual label was activated, and judged that the better the activation, the more efficient the network learning is.

\noindent\textbf{Accuracy and Computational Load Trade-off}
There is a trade-off between accuracy and computational load. For good performance, the network needs to grow, and as the network grows, so does the computational load. Therefore, the goal was to obtain a large increase in accuracy through a lower increase in the computational load.

\subsection{Prerequisite Experiment}
\begin{figure}
\centering     %%% not \center
\subfigure[Using the starting frame as an input]{\label{6.1}\includegraphics[width=0.49\textwidth]{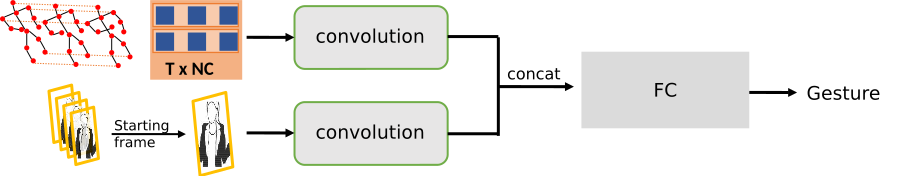}}
\subfigure[Using the keyframe as an input]{\label{6.2}\includegraphics[width=0.49\textwidth]{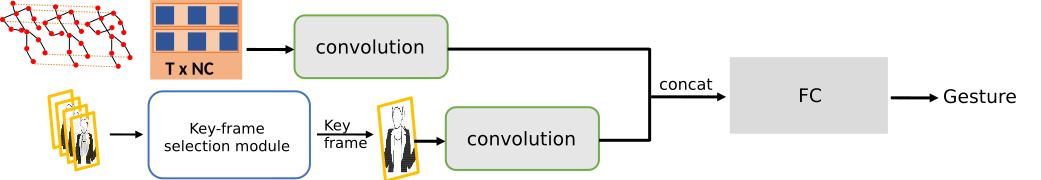}}
\caption{An overview of prerequisite experiment}
\label{6}
\end{figure}

First, we experimented by diversifying the video frames used to classify gestures to prove that keyframes drawn using skeleton data can be used as keyframes for videos. The experiment we designed is shown in Fig. \ref{6}. This experiment identifies changes of the gesture recognition performance depending on input video frames.

In Fig. \ref{6.1}, the input takes the skeleton sequence and the starting frame of the image sequence. The skeleton sequence extends the time dimension and the channel dimension widely to form a single frame, the features of which are extracted through the spatial attention module and temporal attention module. These extracted skeleton features and extracted features from the starting frame are fused and judged through a fully connected layer. By comparison, in Fig. \ref{6.2}, instead of the starting frame of the image sequence, the keyframe of the video drawn through the keyframe selection module is introduced. The skeleton features and extracted features from the keyframe are fused and judged. The performance difference between the two experiments can prove that keyframes drawn using skeleton data can be used as keyframes of videos.

\noindent\textbf{Impact of skeleton-based keyframe selection module}
Table \ref{table:Prerequisite} represents the measurements of accuracy of the underlying network using only skeleton sequences and that of putting the starting frame and keyframe together as an input. Compared to the results tested using only skeleton sequences, we can see a significant increase in performance when image frames are put together. This shows that recognizing gestures with only \textit{x}-axis, \textit{y}-axis, and \textit{z}-axis values of each joint is insufficient in determining similar behaviors. It is necessary to use the spatial features in the image to achieve better results. Furthermore, the results of the difference in frame input show that the keyframe drawn using skeleton data plays an important role. This means that the keyframe spatial feature contains more important information in recognizing gestures than the spatial feature held by a typical starting frame. It is advantageous to use the skeleton-based keyframe selection module, which can achieve outstanding performance.

\begin{table}[t]
\renewcommand*{\arraystretch}{1.4}
\begin{center}
\resizebox{0.45\textwidth}{!}{
\begin{tabular} {ccccccccccc}
\hline\hline
& Network && Accuracy ($\%$) && Difference ($\%p$) &\\
\hline
& Skeleton only && 64.72 && +0 &\\
& Skeleton + starting frame && 67.44 && +2.72 &\\
& \textbf{Skeleton + keyframe} && \textbf{69.65} && \textbf{+4.93} &\\
\hline\hline
\end{tabular}}
\end{center}
\caption{Recognition accuracy from a prerequisite experiment}
\label{table:Prerequisite}
\end{table}

\subsection{Fusion Method}
We experimented by diversifying our method of utilizing skeleton data according to two methods: 1) After constructing the skeleton's frame using skeleton data in the same way as Section 4.1, we proceeded with gesture recognition by drawing features from those frames and concatenating them with features in the video sequence. 2) We construct the first pathway using the skeleton-based keyframe selection module and then proceed with gesture recognition.

\begin{figure}
\centering     %%% not \center
\subfigure[Utilizing skeleton features in a separate pathway]{\label{7.1}\includegraphics[width=0.49\textwidth]{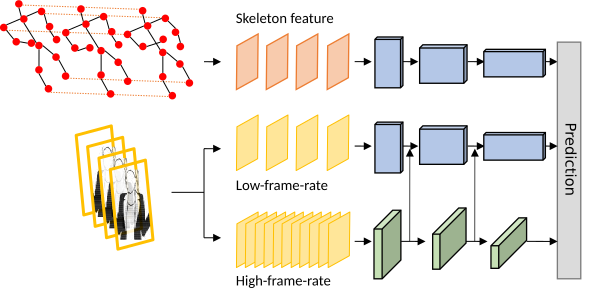}}
\subfigure[Utilizing skeleton data as a keyframe selector]{\label{7.2}\includegraphics[width=0.49\textwidth]{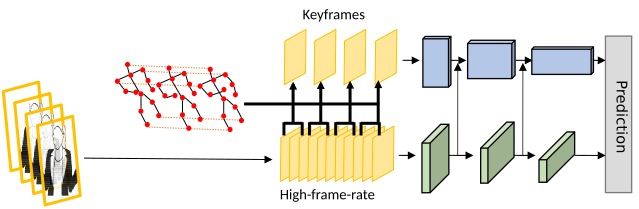}}
\caption{Experiments to determine the influence of fusion methods}
\label{7}
\end{figure}

As can be found in Fig. \ref{7.1}, for the first method, skeleton sequences, high-frame-rate image sequences, and low-frame-rate image sequences enter the network to form three pathways. On the other hand, in Fig. \ref{7.2}, the network consists of two pathways, with a high-frame-rate image sequence and a keyframe image sequence entering the network. The performance difference between the two experiments allows us to find more effective fusion methods.

We measured gesture recognition accuracy to evaluate the performance of each method. We fine-tuned our network on each dataset and compared it to the following networks.

\begin{itemize}
\item C2D \cite{he2016deep}: This is a ResNet-50 frame-based model. We took a pretrained model from ImageNet and proceeded with the learning. Three-dimensional video data was processed using pooling in the time axis direction.
\item I3D \cite{carreira2017quo}: Similarly, we used a pretrained model from ImageNet, copying and pasting the parameters for the time axis three times, but scaling parameter values to $1 \over 3$.
\item Slowfast \cite{feichtenhofer2019slowfast}: This is a two-stream, state-of-the-art network. We exploited the high-frame-rate image sequence and the low-frame-rate image sequence for gesture recognition.
\end{itemize}

We tested the above networks, and we conducted experiments using the slowfast network as the base network for exploiting skeleton data. Experimental results from each fusion method demonstrate that it is better to select keyframes in a video from skeleton data than to draw features from the skeleton data directly.

\noindent\textbf{Effect of varying fusion methods}
Table \ref{2_t} represents the results of testing the above networks. The C2D, I3D, and slowfast networks all used video sequences only. We also ran tests by adding skeleton features to the slowfast network and using the skeleton-based keyframe selection module. Compared to C2D and I3D models, we can confirm that slowfast networks perform very well because they extract spatial and temporal properties effectively. In the low-frame-rate pathway, they extract the spatial features for each frame. In the high-frame-rate pathway, they extract the temporal features for the video sequences. By separating and extracting each characteristic, the network achieves good results as it prevents different features from containing overlapping information.

\begin{table}[t]
\renewcommand*{\arraystretch}{1.4}
\begin{center}
\resizebox{0.45\textwidth}{!}{
\begin{tabular} {ccccccccccc}
\hline\hline
& Network && Accuracy ($\%$) && Difference ($\%p$) &\\
\hline
& C2D \cite{he2016deep} && 80.338 && -5.011 &\\
& I3D \cite{carreira2017quo} && 82.673 && -2.676 &\\
& Slowfast \cite{feichtenhofer2019slowfast} && 85.349 && +0 &\\
& Slowfast + skeleton feature && 85.354 && +0.005 &\\
& \textbf{Slowfast + keyframe selection module} && \textbf{86.151} && \textbf{+0.802} &\\
\hline\hline
\end{tabular}}
\end{center}
\caption{Recognition accuracy according to fusion methods}
\label{2_t}
\end{table}

The method of adding a new pathway using skeleton features did not show a significant increase in performance. This is due to the information that skeleton features have and the fact that video sequences overlap each other. Because skeleton data itself is obtained through a video sequence, it is difficult for skeleton features to contain other important information, except for the information that video data has. However, we can confirm that when skeleton pathway is added, the network outperforms the existing performance by itself.

When keyframes are selected using skeleton sequences, they achieve better performance than conventional networks. This implies that the network has well-selected frames that are effective in recognizing gestures compared to conventional networks and delivers the frames' spatial features effectively. However, there is an improvement point at which the high-frame-rate pathway can decide where to look to obtain efficient temporal features.

\subsection{Gesture Recognition}
We proposed BCCN for effective gesture recognition and conducted comparative experiments with the state-of-the-art networks. We checked the recognition accuracy of the gestures and the activation map of the layer that exists at the end of the network. The gesture recognition accuracy was tested by fine-tuning our network on each dataset, just like previous experiments.

We used GradCAM \cite{selvaraju2017grad} to check the activation map of the layer. If the activated part overlapped with the part needed to determine the actual behavior, it was judged as a good activation, and if not, as a bad activation. Since activation allows us to determine how efficiently the network has been learned, it is important to obtain good activation for good gesture recognition.

\noindent\textbf{Impact of the proposed network}
Table \ref{3_t} represents the results of testing with existing networks and the proposed network. The proposed network performed better compared to the slowfast network, which was previously considered state-of-the-art, and also performed better compared to other networks. This means that the keyframe drawn by the BCCN contained more information, and that information was well communicated over the network. We obtained better results in most classes, especially when classifying small motions.

\begin{table}[t]
\renewcommand*{\arraystretch}{1.4}
\begin{center}
\resizebox{0.45\textwidth}{!}{
\begin{tabular} {ccccccccccc}
\hline\hline
& Network && Accuracy ($\%$) &\\
\hline
& C2D && 80.338 &\\
& I3D && 82.673 &\\
& Slowfast && 85.349 &\\
& Slowfast + skeleton features && 85.354 &\\
& \textbf{Slowfast + keyframe selection module}&& \textbf{86.151} &\\
& \textbf{Ours (BCCN)} && \textbf{87.775} &\\
\hline\hline
\end{tabular}}
\end{center}
\caption{Comparison of recognition accuracy for different networks}
\label{3_t}
\end{table}

\begin{figure}[t]
    \centerline{\includegraphics[width=0.30\textwidth]{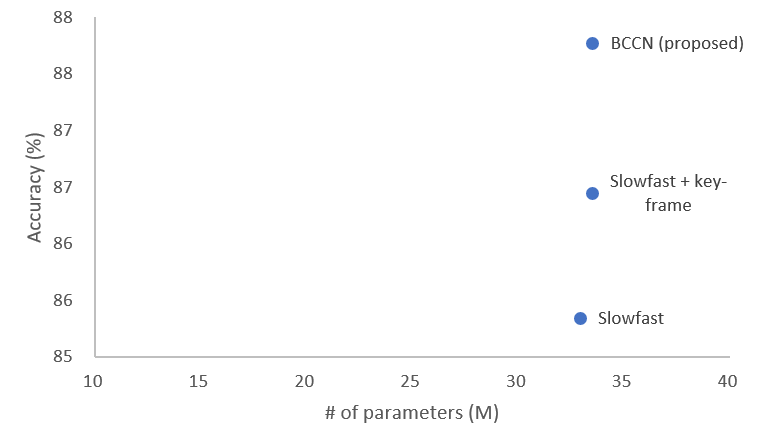}}
    \caption{Computational load analysis for the networks
    } \label{4_t}
\end{figure}

The BCCN was able to achieve good performance without significantly increasing the number of parameters. As shown in Fig. \ref{4_t}, the BCCN outperformed the existing network with only a small increase in the number of parameters, compared to the structure of the existing networks. We confirm that our method is relatively superior in the trade-off between accuracy and computational load.

As shown in Table \ref{5_t}, we tested the effectiveness of the three units comprising the BCCN: the skeleton-based keyframe selection module, the KTT unit, and the TTK unit. As previously demonstrated in several experiments, when we added the skeleton-based keyframe selection module, we could see that the network was good at extracting important spatial features. However, this keyframe feature alone did not effectively capture the temporal semantics. The use of spatiotemporal features that can be obtained from each pathway through KTT and TTK units has become an important factor in gesture recognition. This means that it is important to extract the spatial features of the keyframe, which in turn is important in extracting spatiotemporal semantics.

\begin{table}[t]
\renewcommand*{\arraystretch}{1.4}
\begin{center}
\resizebox{0.45\textwidth}{!}{
\begin{tabular} {ccccccccccc}
\hline\hline
\multirow{2}{*}{Network} & \multirow{2}{*}{Keyframe selection} & \multirow{2}{*}{KTT unit} & \multirow{2}{*}{TTK unit} & \multicolumn{3}{c}{Higher is better} \\
                         &                                     &                           &                           & Accuracy(\%)  & Spatial  & Temporal\\
\hline
Slowfast & & & & 85.349 & Bad & Common\\
Ours & $\bigcirc$ & & & 82.995 & Common & Bad\\
Ours & $\bigcirc$ & & $\bigcirc$ & 86.224 & Common & Common\\
Ours & $\bigcirc$ & $\bigcirc$ & & 86.356 & Common & Good\\
\textbf{Ours (BCCN)} & \textbf{$\bigcirc$} & \textbf{$\bigcirc$} & \textbf{$\bigcirc$} & \textbf{87.775} & \textbf{Good} & \textbf{Good}\\

\hline\hline
\end{tabular}}
\end{center}
\caption{Ablation study}
\label{5_t}
\end{table}

\begin{figure*}[t]
\centering     %%% not \center
\subfigure[Slowfast network]{\label{8.1}\includegraphics[width=.33\textwidth, height=16cm]{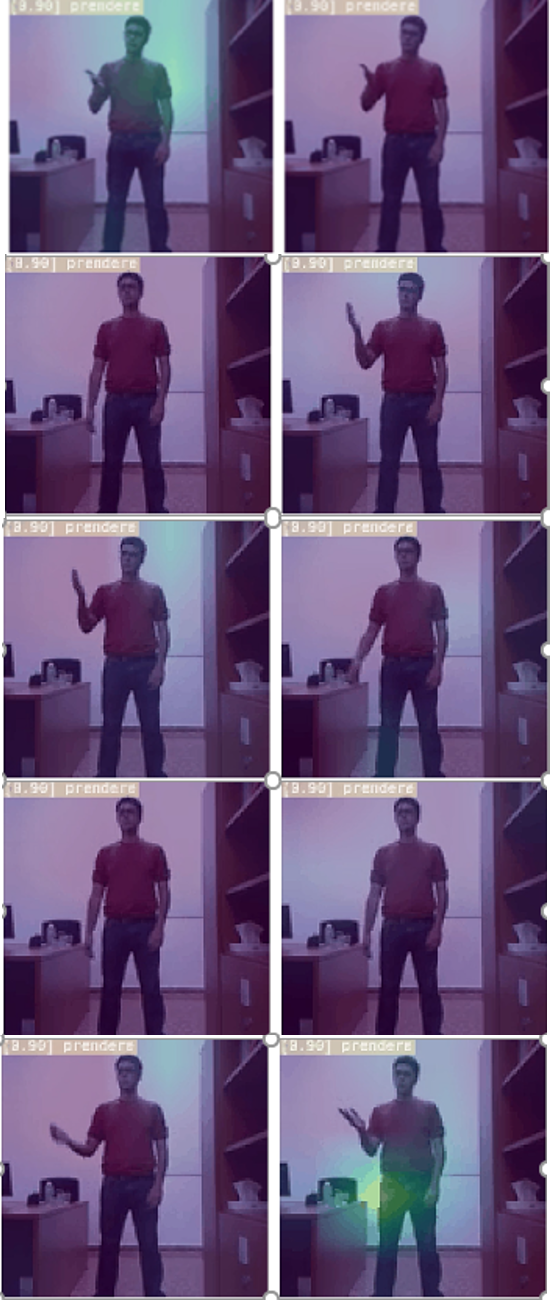}}
\subfigure[Slowfast network + keyframe selection module]{\label{8.2}\includegraphics[width=.33\textwidth, height=16cm]{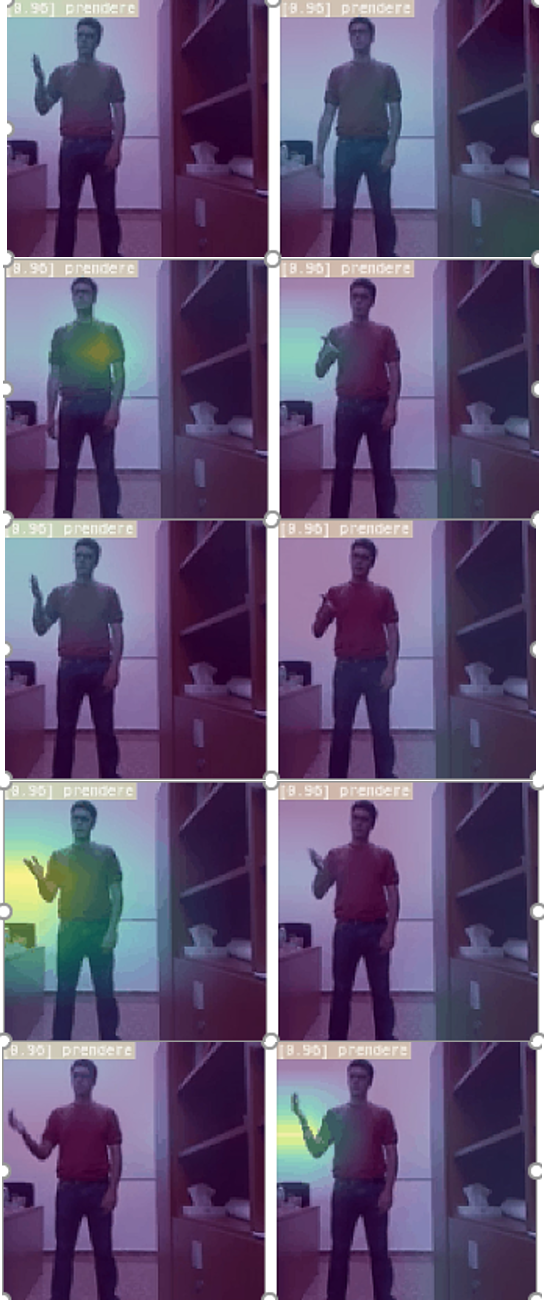}}
\subfigure[BCCN]{\label{8.3}\includegraphics[width=.33\textwidth, height=16cm]{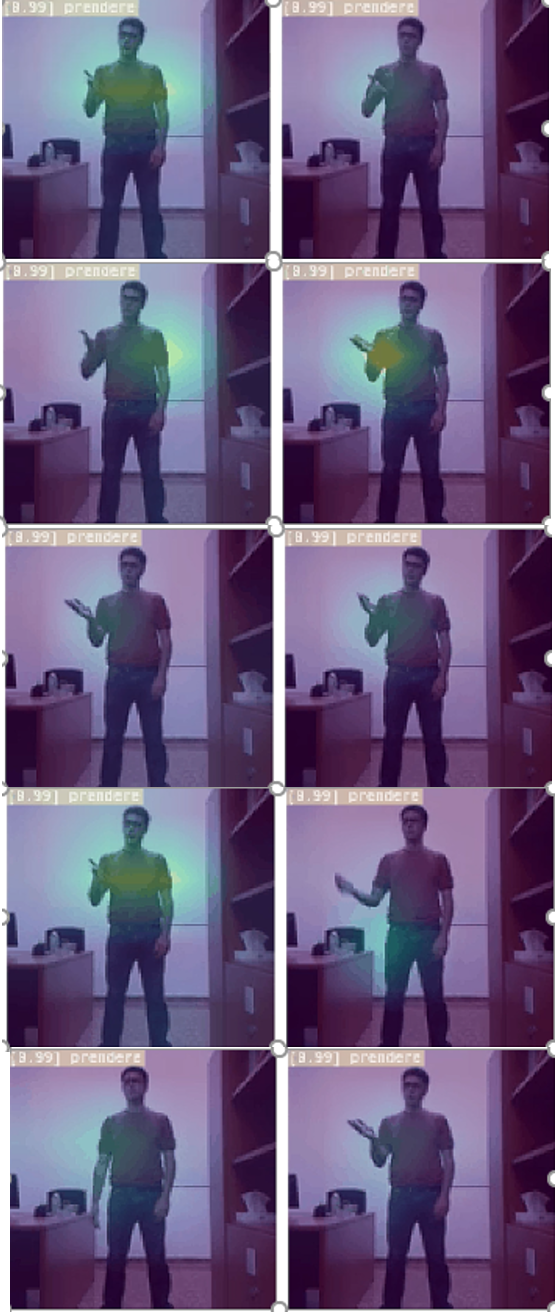}}
\caption{The activation map from each network for each pathway.}
\label{8}
\end{figure*}

The left part of each figure shown in Fig. \ref{8} is the activation map for pathways containing keyframes, and the right part of each figure is the activation map for temporal-attention pathways. The keyframe activation map is relatively well activated for keyframe spatial features and has much information about where to look in recognizing human gestures. The activation map of the temporal-attention pathway is relatively well enabled for temporal semantics. It can be seen as a significant activation at important timing for gesture recognition as well as temporal information within the frame. Fig. \ref{8.1} is an activation map on the base slowfast network, and shows that neither spatial nor temporal activation maps catch hand movements well. Fig. \ref{8.2} shows that when applying a keyframe selection module, it focuses better on hand movements compared to the base network. In addition, the temporal activation map can also be effective when hand movements are important. The BCCN in Fig. \ref{8.3}, was successfully activated on both spatial and temporal activation maps compared to other networks. This means that the BCCN successfully learned the necessary information in recognizing gestures, and has been efficiently trained to focus on important parts. In particular, when looking at temporal activation maps, we can see that spatiotemporal features are well-activated, helping to recognize gestures. Even when obtaining temporal information, the BCCN could focus on the spatial parts and get better temporal features.

\section{Conclusion}
We introduced a method to leverage skeleton data for effective gesture recognition. The skeleton-based keyframe selection module is applicable to all networks, and with a small increase in computational load, it can achieve a superior performance increase over the skeleton feature alone. In addition, we proposed BCCN, a novel network that can effectively convey features of the keyframe extracted through the skeleton-based keyframe selection module. The KTT unit, which can effectively preserve the keyframe feature in the network, and the TTK unit, which allows the spatiotemporal features become insensitive to their size of the feature, were able to learn the features efficiently. Both units allow the two pathways that make up the BCCN to obtain better information from each other. As a future work, we plan to expand the modalities. Each modality has different characteristics effects for extracting features efficiently. Focusing on specific features through networks could help us perceive features more thoroughly, similar to human perception.

{\small
\bibliographystyle{ieee_fullname}
\bibliography{egbib}

\begin{thebibliography}{10}\itemsep=-1pt

\bibitem{argyriou2008convex}
Andreas Argyriou, Theodoros Evgeniou, and Massimiliano Pontil.
\newblock {Convex multi-task feature learning}.
\newblock {\em Machine Learning}, 73(3):243--272, 2008.

\bibitem{cao2019openpose}
Zhe Cao, Gines Hidalgo, Tomas Simon, Shih-En Wei, and Yaser Sheikh.
\newblock {OpenPose: Realtime multi-person 2D pose estimation using part
  affinity fields}.
\newblock {\em TPAMI}, 43(1):172--186, 2019.

\bibitem{carreira2017quo}
Joao Carreira and Andrew Zisserman.
\newblock {Quo vadis, action recognition? a New model and the kinetics
  dataset}.
\newblock In {\em CVPR}, pages 6299--6308, 2017.

\bibitem{caruana1997multitask}
Rich Caruana.
\newblock {Multitask learning}.
\newblock {\em Machine Learning}, 28(1):41--75, 1997.

\bibitem{crawshaw2020multi}
Michael Crawshaw.
\newblock {Multi-task learning with deep neural networks: A survey}.
\newblock {\em arXiv preprint arXiv:2009.09796}, 2020.

\bibitem{das2019toyota}
Srijan Das, Rui Dai, Michal Koperski, Luca Minciullo, Lorenzo Garattoni,
  Francois Bremond, and Gianpiero Francesca.
\newblock {Toyota Smarthome: Real-world activities of daily living}.
\newblock In {\em ICCV}, pages 833--842, 2019.

\bibitem{deng2009imagenet}
Jia Deng, Wei Dong, Richard Socher, Li-Jia Li, Kai Li, and Li Fei-Fei.
\newblock {ImageNet: A large-scale hierarchical image database}.
\newblock In {\em CVPR}, pages 248--255, 2009.

\bibitem{escalera2014chalearn}
Sergio Escalera, Xavier Bar{\'o}, Jordi Gonzalez, Miguel~A Bautista, Meysam
  Madadi, Miguel Reyes, V{\'\i}ctor Ponce-L{\'o}pez, Hugo~J Escalante, Jamie
  Shotton, and Isabelle Guyon.
\newblock {Chalearn looking at people challenge 2014: Dataset and results}.
\newblock In {\em ECCV}, pages 459--473, 2014.

\bibitem{fang2017rmpe}
Hao-Shu Fang, Shuqin Xie, Yu-Wing Tai, and Cewu Lu.
\newblock {RMPE: Regional multi-person pose estimation}.
\newblock In {\em ICCV}, pages 2334--2343, 2017.

\bibitem{feichtenhofer2019slowfast}
Christoph Feichtenhofer, Haoqi Fan, Jitendra Malik, and Kaiming He.
\newblock {SlowFast networks for video recognition}.
\newblock In {\em ICCV}, pages 6202--6211, 2019.

\bibitem{feichtenhofer2017spatiotemporal}
Christoph Feichtenhofer, Axel Pinz, and Richard~P Wildes.
\newblock {Spatiotemporal multiplier networks for video action recognition}.
\newblock In {\em CVPR}, pages 4768--4777, 2017.

\bibitem{feichtenhofer2016convolutional}
Christoph Feichtenhofer, Axel Pinz, and Andrew Zisserman.
\newblock {Convolutional two-wtream network fusion for video action
  recognition}.
\newblock In {\em CVPR}, pages 1933--1941, 2016.

\bibitem{gadzicki2020early}
Konrad Gadzicki, Razieh Khamsehashari, and Christoph Zetzsche.
\newblock {Early vs late fusion in multimodal convolutional neural networks}.
\newblock In {\em Fusion}, pages 1--6, 2020.

\bibitem{gao2020listen}
Ruohan Gao, Tae-Hyun Oh, Kristen Grauman, and Lorenzo Torresani.
\newblock {Listen to look: Action recognition by previewing audio}.
\newblock In {\em CVPR}, pages 10457--10467, 2020.

\bibitem{han2016multi}
Lei Han and Yu Zhang.
\newblock {Multi-stage multi-task learning with reduced rank}.
\newblock In {\em AAAI}, volume~30, 2016.

\bibitem{he2016deep}
Kaiming He, Xiangyu Zhang, Shaoqing Ren, and Jian Sun.
\newblock {Deep residual learning for image recognition}.
\newblock In {\em CVPR}, pages 770--778, 2016.

\bibitem{jang2020etri}
Jinhyeok Jang, Dohyung Kim, Cheonshu Park, Minsu Jang, Jaeyeon Lee, and Jaehong
  Kim.
\newblock {ETRI-Activity3D: A large-scale RGB-D dataset for robots to recognize
  daily activities of the elderly}.
\newblock {\em arXiv preprint arXiv:2003.01920}, 2020.

\bibitem{kim2016weighted}
Hanguen Kim, Sangwon Lee, Youngjae Kim, Serin Lee, Dongsung Lee, Jinsun Ju, and
  Hyun Myung.
\newblock {Weighted joint-based human behavior recognition algorithm using only
  depth information for low-cost intelligent video-surveillance system}.
\newblock {\em Expert Systems with Applications}, 45:131--141, 2016.

\bibitem{kim2015real}
Hanguen Kim, Sangwon Lee, Dongsung Lee, Soonmin Choi, Jinsun Ju, and Hyun
  Myung.
\newblock {Real-time human pose estimation and gesture recognition from depth
  images using superpixels and SVM classifier}.
\newblock {\em Sensors}, 15(6):12410--12427, 2015.

\bibitem{kumar2012learning}
Abhishek Kumar and Hal Daume~III.
\newblock {Learning task grouping and overlap in multi-task learning}.
\newblock {\em arXiv preprint arXiv:1206.6417}, 2012.

\bibitem{liu2015multi}
Wu Liu, Tao Mei, Yongdong Zhang, Cherry Che, and Jiebo Luo.
\newblock {Multi-task deep visual-semantic embedding for video thumbnail
  selection}.
\newblock In {\em CVPR}, pages 3707--3715, 2015.

\bibitem{luvizon20182d}
Diogo~C Luvizon, David Picard, and Hedi Tabia.
\newblock {2D/3D pose estimation and action recognition using multitask deep
  learning}.
\newblock In {\em CVPR}, pages 5137--5146, 2018.

\bibitem{meyerson2017beyond}
Elliot Meyerson and Risto Miikkulainen.
\newblock {Beyond shared hierarchies: Deep multitask learning through soft
  layer ordering}.
\newblock {\em arXiv preprint arXiv:1711.00108}, 2017.

\bibitem{miao2017multimodal}
Qiguang Miao, Yunan Li, Wanli Ouyang, Zhenxin Ma, Xin Xu, Weikang Shi, and
  Xiaochun Cao.
\newblock {Multimodal gesture recognition based on the ResC3D network}.
\newblock In {\em ICCV Workshops}, pages 3047--3055, 2017.

\bibitem{neverova2015moddrop}
Natalia Neverova, Christian Wolf, Graham Taylor, and Florian Nebout.
\newblock {ModDrop: Adaptive multi-modal gesture recognition}.
\newblock {\em TPAMI}, 38(8):1692--1706, 2015.

\bibitem{nishida2015multimodal}
Noriki Nishida and Hideki Nakayama.
\newblock {Multimodal gesture recognition using multi-stream recurrent neural
  network}.
\newblock In {\em Image and Video Technology}, pages 682--694, 2015.

\bibitem{ruder2017overview}
Sebastian Ruder.
\newblock {An overview of multi-task learning in deep neural networks}.
\newblock {\em arXiv preprint arXiv:1706.05098}, 2017.

\bibitem{selvaraju2017grad}
Ramprasaath~R Selvaraju, Michael Cogswell, Abhishek Das, Ramakrishna Vedantam,
  Devi Parikh, and Dhruv Batra.
\newblock {Grad-CAM: Visual explanations from deep networks via gradient-based
  localization}.
\newblock In {\em ICCV}, pages 618--626, 2017.

\bibitem{shi2019skeleton}
Lei Shi, Yifan Zhang, Jian Cheng, and Hanqing Lu.
\newblock {Skeleton-based action recognition with directed graph neural
  networks}.
\newblock In {\em CVPR}, pages 7912--7921, 2019.

\bibitem{snoek2005early}
Cees~GM Snoek, Marcel Worring, and Arnold~WM Smeulders.
\newblock {Early versus late fusion in semantic video analysis}.
\newblock In {\em ACM MM}, pages 399--402, 2005.

\bibitem{song2020gesture}
Xiaoyu Song, Hong Chen, and Qing Wang.
\newblock {A gesture recognition approach using multimodal neural network}.
\newblock In {\em Journal of Physics: Conference Series}, volume 1544, page
  012127. IOP Publishing, 2020.

\bibitem{taylor2010convolutional}
Graham~W Taylor, Rob Fergus, Yann LeCun, and Christoph Bregler.
\newblock {Convolutional learning of spatio-temporal features}.
\newblock In {\em ECCV}, pages 140--153, 2010.

\bibitem{toshev2014deeppose}
Alexander Toshev and Christian Szegedy.
\newblock {DeepPose: Human pose estimation via deep neural networks}.
\newblock In {\em CVPR}, pages 1653--1660, 2014.

\bibitem{tran2015learning}
Du Tran, Lubomir Bourdev, Rob Fergus, Lorenzo Torresani, and Manohar Paluri.
\newblock {Learning spatiotemporal features with 3D convolutional networks}.
\newblock In {\em ICCV}, pages 4489--4497, 2015.

\bibitem{wang2020deep}
Yikai Wang, Wenbing Huang, Fuchun Sun, Tingyang Xu, Yu Rong, and Junzhou Huang.
\newblock {Deep multimodal fusion by channel exchanging}.
\newblock {\em arXiv preprint arXiv:2011.05005}, 2020.

\bibitem{xiu2018pose}
Yuliang Xiu, Jiefeng Li, Haoyu Wang, Yinghong Fang, and Cewu Lu.
\newblock {Pose Flow: Efficient online pose tracking}.
\newblock {\em arXiv preprint arXiv:1802.00977}, 2018.

\bibitem{zhu2017multimodal}
Guangming Zhu, Liang Zhang, Peiyi Shen, and Juan Song.
\newblock {Multimodal gesture recognition using 3-D convolution and
  convolutional LSTM}.
\newblock {\em IEEE Access}, 5:4517--4524, 2017.

\end{thebibliography}
}

\end{document}